\documentclass[journal]{IEEEtai}

\usepackage[colorlinks,urlcolor=blue,linkcolor=blue,citecolor=blue]{hyperref}

\usepackage{color,array}

\usepackage{graphicx}

\usepackage{graphicx}
\graphicspath{{./figures/}}
\usepackage{amsmath}
\usepackage{amssymb}
\usepackage{booktabs}
\usepackage{hyperref}
\usepackage{multirow}
\usepackage[colorinlistoftodos]{todonotes}
\usepackage{orcidlink}

\begin{document}
	
	\title{Multiscale and Multilayer Contrastive Learning for Domain Generalization} 

	\author{Aristotelis Ballas \orcidlink{0000-0003-1683-8433} and Christos Diou \orcidlink{0000-0002-2461-1928}, \IEEEmembership{Member, IEEE}
		\thanks{Manuscript created February, 2023; revised 27 May 2023, 27 August 2023, 27 November 2023 and 30 January 2024; accepted 8 March 2024. The work leading to these results has received funding from the European Union’s Horizon 2020 research and innovation programme under Grant Agreement No. 965231, project REBECCA (REsearch on BrEast Cancer induced chronic conditions supported by Causal Analysis of multi-source data).}
		\thanks{A. Ballas is with the Department of Informatics and Telematics, Harokopio University of Athens, Tavros, 177 78 Greece (e-mail: aballas@hua.gr).}
		\thanks{C. Diou is with the Department of Informatics and Telematics, Harokopio University of Athens, Tavros, 177 78 Greece (e-mail: cdiou@hua.gr).}
		}
	
	\markboth{MANUSCRIPT ACCEPTED IN IEEE TRANSACTIONS ON ARTIFICIAL INTELLIGENCE, March 2024}
	{Ballas \MakeLowercase{\textit{et al.}}: $M^2$-CL for Domain Generalization}
	
	\maketitle
	
	\begin{abstract}
		During the past decade, deep neural networks have led to fast-paced progress
		and significant achievements in computer vision problems, for both academia
		and industry. Yet despite their success, state-of-the-art image
		classification approaches fail to generalize well in previously unseen
		visual contexts, as required by many real-world applications. In this paper,
		we focus on this domain generalization (DG) problem and argue that the
		generalization ability of deep convolutional neural networks can be
		improved by taking advantage of multi-layer and multi-scaled representations
		of the network. We introduce a framework that aims at improving domain
		generalization of image classifiers by combining both low-level and
		high-level features at multiple scales, enabling the network to implicitly disentangle representations in its latent space and learn domain-invariant attributes of the depicted objects. Additionally, to further facilitate robust representation learning, we propose a novel objective function, inspired by contrastive learning, which aims at constraining the extracted representations to remain invariant under distribution shifts. We demonstrate the effectiveness of our method by evaluating on the domain generalization datasets of PACS, VLCS, Office-Home and NICO. Through extensive experimentation, we show that our model is able to surpass the performance of previous DG methods and consistently produce competitive and state-of-the-art results in all datasets\footnote{Code available at: \href{https://github.com/aristotelisballas/m2cl}{https://github.com/aristotelisballas/m2cl}}.
	\end{abstract}
	
	\begin{IEEEImpStatement}
		Domain Generalization is one of the most important problems in machine learning today. Popular image classification architectures show significant performance degradation when evaluated on data originating from different distributions than the one(s) they were trained on. This is a common scenario in real-world applications, rendering it a significant constraint of current state of the art image classification models. The domain generalization algorithm we introduce in this paper attempts to overcome such limitations by utilizing and regularizing representations across convolutional neural network architectures, in order to learn domain-invariant attributes of an image. With a noteworthy increase in model accuracy, our algorithm is able to set the state-of-the-art in a total of 4 widely accepted DG datasets. In addition, qualitative samples of our model's inference process provide further validation of our claims\footnote{Paper \href{https://ieeexplore.ieee.org/document/10472869}{DOI}: {10.1109/TAI.2024.3377173}}.
	\end{IEEEImpStatement}
	
	
	\vspace{10mm}
	\vspace{1.2cm}
	\section{Introduction}
	
	\IEEEPARstart{H}{uman} beings are capable of creating abstractions in order to better understand their environment. Even from an early age, we are able to easily extract knowledge from different settings and generalize it to previously unknown circumstances. When classifying images, we base our decisions on the distinguishable characteristics of the depicted class and stand unfazed when presented with irregular qualities -seeing a cat orbiting the moon may seem odd, but we still recognize it as a cat-. In contrast, state-of-the-art computer vision models often fail to mimic a human's adaptability prowess and perform very poorly when presented with previously `unseen' contexts. 
	
	In recent years, Deep Learning models have reached, or even surpassed, human
	accuracy on several difficult computer vision tasks
	\cite{he2015delving}. Despite their outstanding success, these models fail to maintain their performance at a high level when presented with data
	originating from a different distribution than the one(s) they were trained on
	\cite{recht2019imagenet, zhou2022domainold}. In particular, popular
	Convolutional Neural Network (CNN) architectures fail to differentiate between
	spurious correlations or data biases (e.g background, position, selection bias,
	etc) and truly class-representative features of an image.
	
	\begin{figure}
		\centering
		\includegraphics[width=\linewidth]{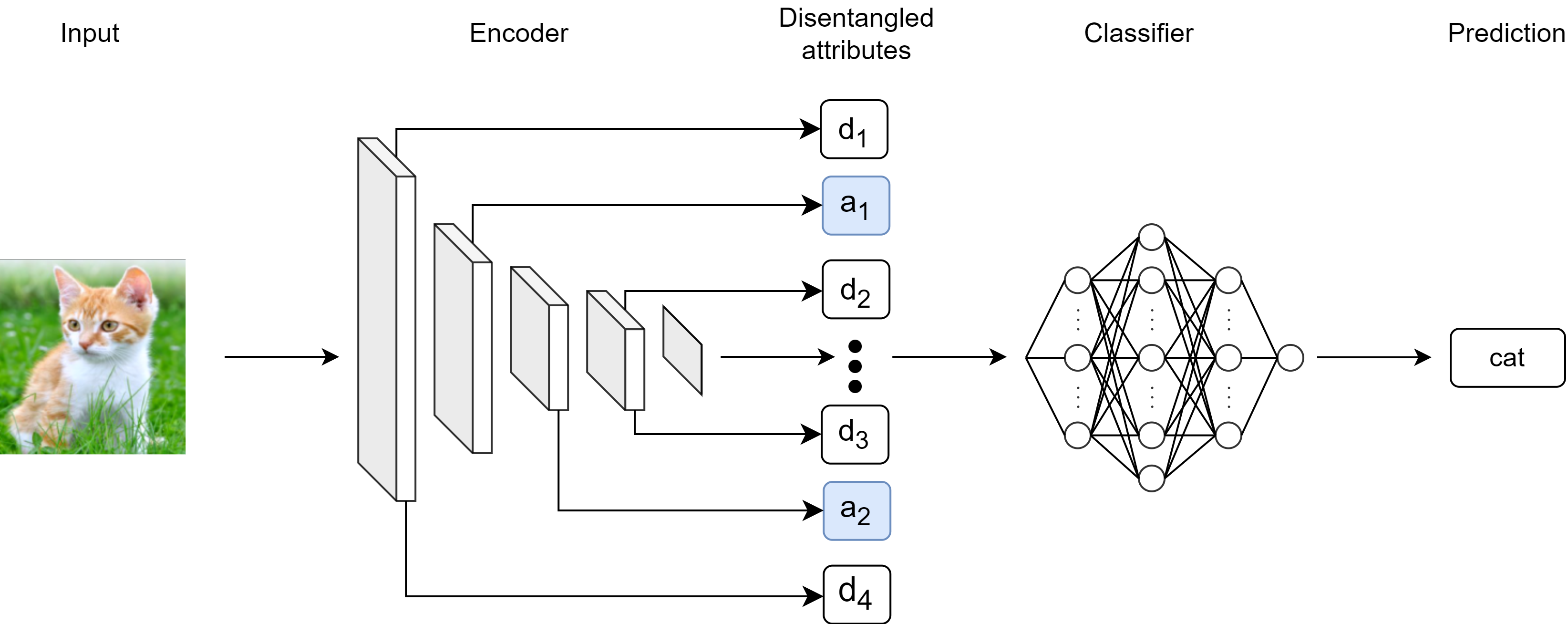}
		\caption{In this work, the main goal is to improve the ability of a model to uncover a \textit{disentangled} representation of an input image. This \textit{disentangled} representation of the image can be thought of as a sequence of class-specific $a_1, a_2$ or domain-specific $d_1, d_2, d_3, d_4$ (but perhaps class-irrelevant) attributes. For example, a domain-specific feature can be attributed to a ``patch of grass'' in the input image, whereas the ``whiskers'' or ``shape of ear'', can be thought of as class-specific. We argue that the problems caused by domain shift between data drawn from unknown domains can be mitigated by utilizing multiple levels of information passed throughout a Convolutional Neural Network, in order to derive disentangled representations. A classifier trained on such disentangled representations can then learn to infer only on the class-specific or causal attributes of the object depicted in the image (blue rectangles).}
		\label{fig:part_based}
	\end{figure}
	
	Several approaches have been proposed to address this
	problem (Section \ref{sec:related_work}). These include methods that aim to extract common semantic features among source domains, emulating biases found in nature, synthesizing samples by mixing data characteristics or using self-supervision to discover invariances in images.
	
	In this paper, we argue that allowing a model to learn by using features of varying complexity and at multiple scales, facilitates the disentanglement of deep representations which tend to incorporate spurious correlations 
	and enables a classifier to learn based on domain-invariant features.
	To this end, we propose a framework which leverages information extracted from intermediate layers of a CNN, combined with a loss function to disentangle the invariant features of the target class. Our method introduces an extraction block which consolidates feature maps from layers across a backbone CNN and uses them for robust, non-identically distributed image classification. To further facilitate the feature disentanglement capabilities of our model, we also propose adding a novel contrastive loss function which explicitly regularizes the training process into extracting causal, class-dependent representations which remain invariant under data distribution shifts. Inspired by the multi-scale and multi-layer representation processing, we name our method $\mathbf{M^2}$. With the addition of the proposed contrastive loss, our method becomes $\mathbf{M^2}$\textbf{-CL}.
	
	We evaluate both of our models on four domain generalization benchmarks, PACS
	\cite{Li_2017_ICCV}, VLCS \cite{5995347}, 
	Office-Home \cite{venkateswara2017deep} and NICO \cite{HE2021107383}. Several
	ablation studies demonstrates the necessity of each component in our proposed
	framework and provide further intuition regarding the effectiveness of the
	proposed approach. While our $\mathbf{M^2}$ model is able to consistently produce competitive results and set the state-of-the-art in several cases, the addition of our proposed novel contrastive loss function boosts its performance in each setting. 
	Our contributions in this work, include the following:
	\begin{itemize}
		\item We introduce a neural network architecture, namely $\mathbf{M^2}$, that uses
		\textit{extraction blocks} and \textit{concentration pipelines} to
		implicitly disentangle representations in the models' latent space and disregards the ones corresponding to spurious correlations between data from distinctive domains.
		\item We propose the addition of a novel training loss function inspired by contrastive learning, for further facilitating the training of our proposed model, called $\mathbf{M^2}$\textbf{-CL}.
		\item We experimentally demonstrate the effectiveness of our method on four widely accepted Domain Generalization benchmarks.
		\item We illustrate, through examples, that our model indeed seems to focus
		on the invariant properties of the objects by producing saliency maps of
		multiple, unseen, input images. These examples indicate that when compared to a baseline model, our framework pays less attention to the contexts present in the image and instead makes predictions based on causal features.
	\end{itemize}

	\section{Previous Work}
	\label{sec:related_work}
	There exist several previous works related to domain generalization in the
	machine learning literature. In this section, we summarize some of the most
	important contributions to the field and briefly present related works regarding robust representation learning algorithms.
	
	\textbf{Domain Adaptation} (DA) \cite{wang2018deep} has made noteworthy progress in the past few
	years. DA algorithms take advantage of pre-trained models and leverage their
	feature extraction capabilities, by fine-tuning them on previously unknown data
	distributions or target domains. To mitigate the distribution shift problem
	between source and target domain data, \cite{li2018domain, pmlr-v28-muandet13, 	pmlr-v70-long17a,NIPS2016_ac627ab1,pmlr-v37-long15} use an MMD-based (Maximum
	Mean Discrepancy) loss, in an attempt to align the source and target
	distributions. For the same purpose, \cite{Chen_Zhao_Liu_Cai_2020,
		ganin2016domain, Tzeng_2017_CVPR} use an adversarial loss. Another popular
	approach in DA, is the identification of invariant representations between
	data. To this end, Generative Adversarial Networks (GAN)
	\cite{NIPS2014_5ca3e9b1} were proposed.  GANs consist of a generator and a
	domain discriminator. Namely, the generator's goal is to synthesize and mimic
	data from the source domains, in order to fool the discriminator. Furthermore,
	in \cite{Li_2021_CVPR_invariant} the authors introduce the LIRR algorithm in
	Semi-Supervised DA, for learning invariant representations and risks. Other
	methods, such as pseudo-labeling \cite{pmlr-v80-xie18c,Chen_2019_CVPR},
	reconstruction \cite{10.1007/978-3-319-46493-0_36,Li_2021_CVPR}, regularization
	\cite{Tompson_2015_CVPR,saito2018adversarial,balaji_metareg_2018}, unsupervised
	domain adaptation with mixup training \cite{yan2020improve} and self-ensembling
	\cite{french2018selfensembling}, also prove to be effective in the DA setting.

	\textbf{Disentangled Representation Learning} \cite{6472238,9363924} aims to
	decompose the input data into separate disentangled independent factors. The
	authors of \cite{9363924} argue that by disentangling their feature space,
	models can leverage the causal attributes of the deconstructed data during
	downstream tasks. A common practice is to train generative models, such as GANs
	or VAEs \cite{kingma2014autoencoding,burgess_understanding_2018}, paired with
	latent space constraints such as mutual information \cite{NIPS2016_7c9d0b1f},
	information bottleneck \cite{9558836} and KL-divergence
	\cite{pmlr-v80-kim18b,Yang_2021_CVPR}.
	
	\textbf{Contrastive Learning} \cite{oord2018representation} is a machine learning technique proposed for extracting useful representations from unlabeled images. As its name suggests, the core principle behind contrastive learning is contrasting images against each other in order to learn image features which remain similar across data classes and representations which separates one class from another.
	
	To this end, a plethora of works utilizing contrastive learning have been proposed. Most noteworthy, SimCLR \cite{chen2020simple} focuses on maximizing the similarity of representations extracted from augmented views of the same image (positive pairs), while also maximizing the dissimilarity of representations extracted from different images (negative pairs). Similarly, MoCO \cite{he2020momentum} utilizes a memory bank for handling a larger number of negative pairs in each batch during training, while other implementations such as BYOL \cite{grill2020bootstrap} and SimSiam \cite{chen2021exploring} do not use negative pairs at all. 
	
	
	Finally, \textbf{fine-grained image analysis} (FGIA) \cite{wei2021fine} algorithms deal with distinguishing visual objects from subordinate categories (e.g different species of dogs). Models tasked with fine-grained image classification attempt to learn representations which correspond to subtle details, such as annotated key point locations, which assist in differentiating between inter-class image variations. To tackle FGIA, the authors of \cite{wang2018learning, ding2019selective} and \cite{huang2020interpretable} proposed training end-to-end fine-grained CNN models. In their work, the authors significantly improved a base models recognition capabilities by employing an additional $1\times1$ convolutional filter for detecting small patches in images. In addition, attention mechanisms \cite{zheng2019looking, ji2020attention} have also been proposed achieving strong accuracy in downstream tasks.
	
	\begin{figure*}
		\centering
		\includegraphics[width=\linewidth]{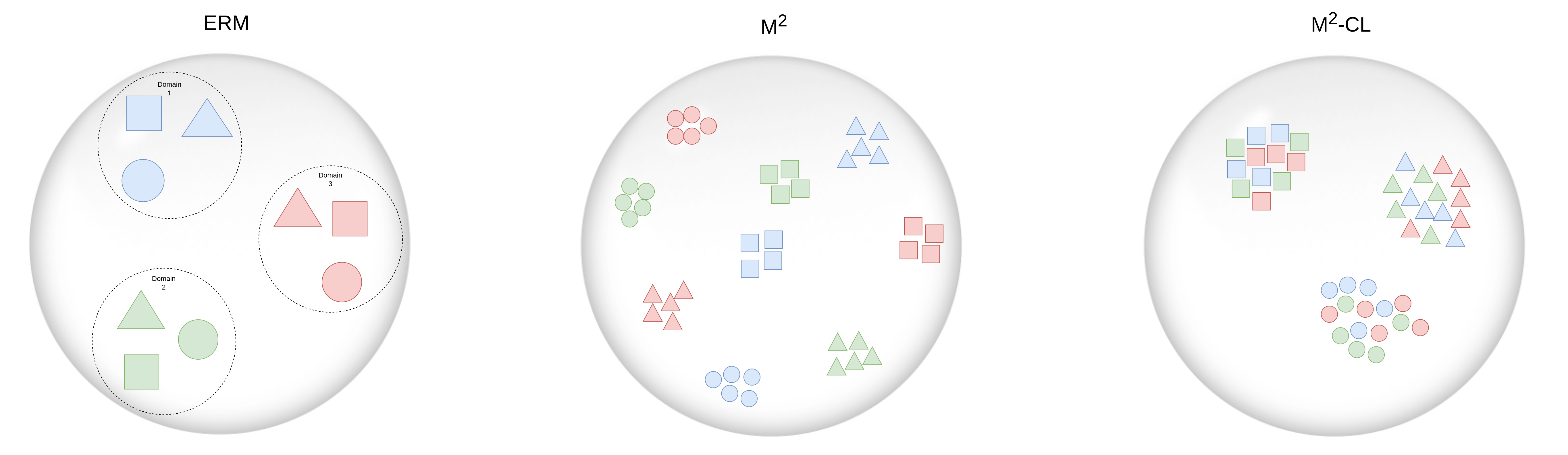}
		\caption{Each object in a certain class consists of distinguishable class-specific attributes which remain invariant between domains, e.g. a cat has whiskers whether there is snow or grass in the background of the image. However, the image also contains domain-specific but note necessarily class-relevant attributes which are entangled in the representation extracted by the final convolutional layers of popular CNN architectures, such as ResNets. The goal in DG is to encode such common class-specific attributes of images, in order to enable models to identify target classes across data domains. Classic fully supervised models learning solely via empirical risk minimization (ERM) tend to correlate the latent space representations with features found in distinct domains. In the above visualization, each representation attributed to a certain class is illustrated with a different shape (circle, square, triangle), while each color (blue, green, red) corresponds to distinct data domains. To mitigate the issues caused by domain shifts in data distributions originating from different data-generating processes, in this work we propose an alternative approach which: a) attempts to derive a set of disentangled representations by extracting multiple levels of information from intermediate CNN layers of the backbone network ($\mathbf{M^2}$) and b) brings the representations extracted from images of the same label closer together, while simultaneously pushes the ones originating from different classes further away ($\mathbf{M^2}$\textbf{-CL}).} 
		\label{fig:method_new}
	\end{figure*} 
	
	\textbf{Domain Generalization} \cite{zhou2022domainold} methods focus on maintaining high accuracy
	across both known and unknown data distributions. The core difference between
	DG and other settings is the fact that during training the model has no access
	to the test domains/distributions whatsoever, which makes it a significantly more difficult problem than DA. Furthermore, in contrast to FGIA algorithms which are tasked to recognize inter-class representation variability, DG aims to employ mechanisms which extract image features that remain invariant under domain shift.
	The DG problem can be split into two settings \cite{zhou2022domainold}: Multi-Source DG and Single-Source DG. In \textit{multi-source DG}, it assumed that data is sampled from multiple distinct but similar domains. Therefore, domain labels are leveraged to learn stable representations across the source domains. In contrast, \textit{single-source DG} assumes that the training data is sampled from a single distribution and is unaware of the presence of separate domains. Our work falls in the second category, as it does not take advantage of domain labels and acts in a domain-agnostic manner. Similarly, the authors of \cite{Zhang_2021_CVPR} use Random Fourier Features and sample weighting to compensate for the complex, non-linear correlations among non-iid data distributions. In \cite{huangRSC2020}, the authors introduce RSC, a self-challenging training heuristic that discards representations associated with the higher gradients. \cite{Carlucci_2019_CVPR} proposes solving jigsaw puzzles via a self-supervision task, in an attempt to restrict semantic feature learning. On the other hand, a plethora of works have been proposed for multi-source DG as well. The authors of \cite{sun2016deep}
	introduce deep CORAL, a method to align correlations of layer activations in
	DNNs. Meta-learning approaches have also been proposed, in \cite{li2018learning,
		pmlr-v70-finn17a} and Adaptive Risk Minimization (ARM)
	\cite{zhang2021adaptive}, which leverage a meta-learning paradigm for adapting
	to unseen domains. Additionally, \cite{10.1007/978-3-030-58607-2_12}
	takes advantage of the variational bounds of mutual information in the meta-learning setting and uses episodic training to extract invariant representations. In another approach, Style-Agnostic networks, or SagNets \cite{nam2021reducing}, attempt to reduce the gap between domains, by focusing on disentangling style encodings. Furthermore, the authors of
	\cite{10.1007/978-3-030-58542-6_5} combine batch and instance normalization to
	extract domain-agnostic feature representations, while \cite{ballas_cnn_2023} and \cite{zhou2021domain} explore the use of data augmentation and style-mixing techniques in the DG setting, respectively. The authors of \cite{eastwood2022probable} propose to address DG by formulating an empirical quantile risk minmization problem (EQRM), in order to learn predictors with high probability. More recently, SAGM \cite{wang2023sharpness} proposes aligning the gradient 
	directions between the empirical risk and a perturbation loss for improved 
	model generalization. Finally, \textit{ensemble} methods seem to yield a 
	significant performance improvement in several DG benchmarks. For example, SIMPLE \cite{li2023simple} leverages a large number of pre-trained models and 
	deploys the most appropriate one for prediction, based on a matching metric
	on target OOD samples. Similarly, SWAD-based methods \cite{cha_swad_2021} have also been proposed. MIRO \cite{avidan_domain_2022} takes advantage of 
	an Oracle model and regularizes pre-trained models via Mutual Information, 
	while PCL \cite{yao_pcl_2022} uses a proxy-based contrastive method in 
	conjuction with SWAD.
	
	For our method, we drew inspiration from previous \textbf{relevant research.}
	The utilization of multi-layer representations has been explored before with several proposed NN architectures. In \cite{Hariharan_2015_CVPR}, the authors introduced the Hypercolumns and Efficient Hypercolums methods in order to take advantage of the different level of information throughout a neural network. However, in order to use the above methods it is expected that bounding boxes for the points of interest of the image are provided. By adapting the Hypercolumns method to the image classification DG setting, the proposed algorithm in \cite{ballasdiou2021} demonstrates improved performance in the NICO dataset over a fully-supervised baseline model. The authors of \cite{ronneberger_u-net_2015}, propose upsampling extracted feature maps from intermediate layers of a CNN and concatenating them to later layers of the network, for biomedical image segmentation. The processing of intermediate features, has also been reported to improve performance in other fields, such as 1D biosignal classification \cite{10233054}. By directly connecting any layer, and passing all its feature maps to all subsequent layers, DenseNets \cite{huang_densely_2017} explore feature reuse for classic image classification downstream tasks. Regularization techniques for learning class-discriminative features have also been proposed in the past. \cite{zhang_be_2019} proposes a self-distillation loss that injects the model's softmax output into the shallower layers to learn more discriminative features of a class object. \cite{wang2021revisiting} shows if highly class-discriminative features are learned in the shallow layers, even the class-relevant information required for the deeper layers may be discarded. Specifically for DG, \cite{kim_selfreg_2021} proposes SelfReg which uses self-supervised contrastive losses coupled with class-specific domain perturbation layers and gradient stabilization techniques to learn domain-invariant features of a class, while the authors of \cite{ballas_cnns_2023} propose applying attention mechanisms to intermediate feature maps. Our framework, in contrast to previously proposed algorithms, encodes the extracted information without injecting it back into the network and potentially propagating spurious, domain-relevant features. Furthermore, the addition of our proposed contrastive loss term in the training process enables our model to extract and encode class-relevant representations at every level of the network, without needing to solely rely on the information extracted from the last layer, as in most papers leveraging contrastive learning frameworks.

	\begin{figure*}[t]
		\centering
		\includegraphics[width=\linewidth]{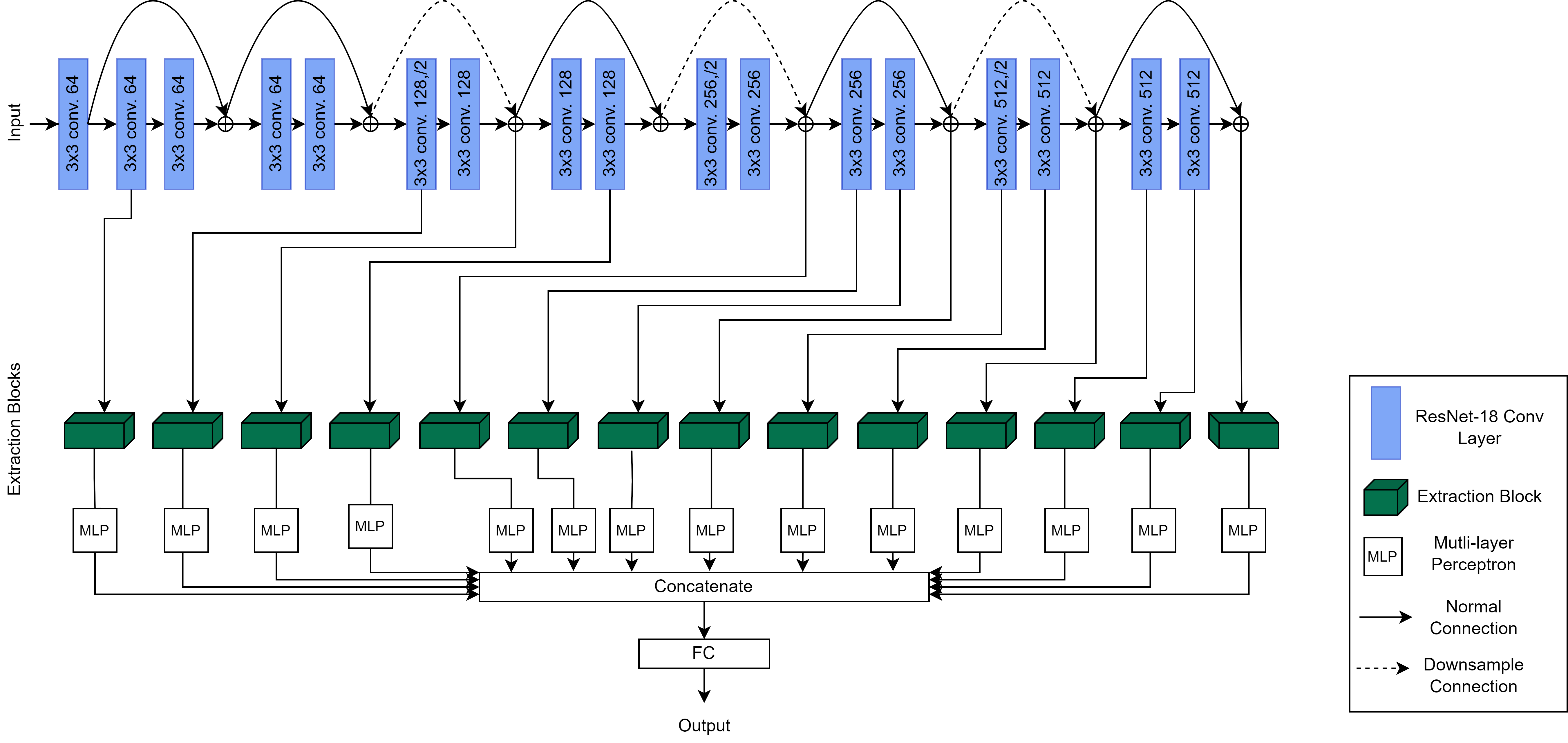}
		\caption{Visualization of the M\textsuperscript{2} architecture built on a ResNet-18. We propose extracting feature maps (black arrows) from layers across the ResNet-18 with the use of multiple \emph{extraction blocks} (green boxes). The lines above the network's conv layers represent ResNet skip connections. The solid lines indicate that the feature maps retain their dimension, while the feature maps passed through the dashed connection lines are downsampled to match the dimension of the previous layer output. Our network's main functionality derives from the multiple parallel \emph{concentration pipelines} in each extraction block. By utilizing feature maps from intermediate outputs of convolutional layers in the backbone model, our framework combines low-level, multi-scale features of early layers with more complex features extracted at layers further down the network. The parallel pipelines, aim at processing the extracted feature maps in a multi-scale manner, each emphasizing on a different characteristic of the object depicted in the input image. We argue that by incorporating outputs from different intermediate levels of the network, we enable the model to disentangle the invariant qualities of an image.}
		\label{fig:model}
	\end{figure*}

	\section{Methods}

	
	\subsection{Motivation - Entangled Representations in Deep CNN Image Classification}
	There exists an intuition as to how deep CNNs learn to classify
	images. Based on illustrations \cite{olah2017feature, olah2018building}
	of the learnt features of trained convolutional filters in a CNN, early layers tend to
	detect low-level features in an image, such as edges or lines, whereas
	the network's deeper layers build upon these features through feature
	composition to encode more complex representations, such as curves, shapes,
	textures and ultimately, objects.
	
	When building such complex representations, feature composition may encode
	class-specific features jointly with domain-specific (but not necessarily class-relevant) attributes, into non-separable representation components.
	In this case, the network may learn to distinguish classes based on domain-specific shortcuts \cite{geirhos2020shortcut} and therefore fail to generalize across domains (i.e., data distributions).
	In fact, as formally shown in \cite{cohen2016expressive},
	\cite{cohen2017inductive} and \cite{levine2018deep}, polynomially-sized deep
	convolutional networks can model complex correlations (exponentially high
	separation ranks) between groups of image regions. Without any remedial measures, if a correlation exists between the training domains and the target classes, it will be encoded in the learnt representation. An intuitive example
	regarding domain specific and class-specific or -irrelevant attributes, is depicted in Fig.\ref{fig:part_based}.
	
	We can express the above notion in a formal manner by following the definition of disentangled group actions and representations, as presented in \cite{higgins2018towards}. To better understand the concept of a disentangled representation, we first describe the properties of a \textit{disentangled group action}\footnote{We use the following notation: group ($G$), group decomposition into a direct product of subgroups $(G = G_1 \times G_2)$, set $X$, group action on $X$ $(\cdot : G \times X \rightarrow X)$, action of the $i$-th subgroup $\cdot_i$.}. Suppose that there exists
	a group $G$, acting on a set $X$, via an action $\cdot : G \times X \rightarrow X$. Furthermore, suppose that $G$ decomposes into the direct product of two
	subgroups, such that $G = G_1 \times G_2$, and that each subgroup can act on $X$ via an action $\cdot_i$, $i \in {1, 2}$. If there exists a
	decomposition $X = X_1 \times X_2$ and for all $x^{(1)}_{1}, x^{(1)}_{2} \in X_1$ and $x^{(2)}_{1}, x^{(2)}_{2} \in X_2$, it holds that
	\begin{equation}
		(x^{(1)}_{1}, x^{(2)}_{1}) \cdot (x^{(1)}_{2}, x^{(2)}_{2}) = (x^{(1)}_{1} \cdot_1 x^{(1)}_{2}, x^{(2)}_{1} \cdot_2 x^{2}_{2})
	\end{equation}
	then we say that the the group action is \textit{disentangled}, with respect to the
	decomposition of $G$. This definition can be directly extended for decompositions into a larger number of groups, i.e., $G = G_1 \times \dotsc \times G_n$. In this case, we say that the group action is disentangled with respect to the decomposition of the group, if there exists a decomposition $X = X_1 \times \dotsc \times X_n$, such that each $X_i$ is invariant to all actions, $\cdot_j$, of the subgroups $G_j, j \neq i$ and is only affected by the action of $G_i$, with $i, j = 1, \dotsc, n$.
	
	
	The above formalization can, in turn, be extended to the internal representations of a model. Suppose we have a set of world-states $W$, a generative process which leads to observations $O$ of the world (e.g., pixels of an image) and a model that processes the observations, leading to a set of internal representations, $Z$.
	Additionally, suppose that there exists a group $G$ acting on $W$ via an action
	$\cdot : G \times W \rightarrow W$, which can be decomposed as a direct product
	$G = G_1 \times \cdots \times G_n$. In our case, we assume that each subgroup action, $\cdot_{i}$ leads to an alteration of different image attributes. For example, one action could change class-related features (e.g., presence and appearance of a cat in an image scene) and a different action alters domain attributes (e.g., presence and appearance of grass).
	The goal of our work is to guide a model to produce a set of \textit{disentangled representations} $Z$ with respect to the decomposition of $G$, such that the set $Z$ can be decomposed into $Z = Z_1 \times \cdots \times Z_n$ and that elements of each $Z_i$ are invariant to the actions of all $G_j, j \neq i$.
	
	In the case of CNN models for image classification, we propose using intermediate layers of the CNN to derive the sets $Z_i$, instead of relying on the final, entangled layer of the model. Specifically, let the set of representations be
	\begin{equation}
		Z = Z_1 \times \cdots \times Z_{L-1}
	\end{equation}
	where each $Z_i$ is a set of (filter) outputs derived from an intermediate hidden layer and $L$ the number of such intermediate outputs. Given an input image, if $\mathbf{z}_i \in Z_i$ is the output of the $i$-th filter, then the model's internal representation is $\mathbf{z} = \begin{bmatrix}\mathbf{z}_1 & \dotsc & \mathbf{z}_L\end{bmatrix}$. Training the model using this representation allows it to encode class-specific and domain-specific attributes separately, before these become entangled into progressively more complex features of subsequent layers.
	
	
	
	Directly implementing this approach, however, is impractical. The outputs of
	intermediate CNN layers are (a) typically very large, leading to dimensionality
	and memory restrictions and (b) contain a lot of redundant, non-discriminative
	information. In this paper, we implement the proposed algorithm and address the aforementioned problems by considering
	only a subset of intermediate layer outputs and subsequently processing the output of
	such layers through \emph{extraction blocks}, before concatenating them to the
	final network representation, with the aim to derive concise and discriminative
	features of the input image.

	\subsection{Overview of the Proposed Approach }
	
	
	The proposed architecture implements extraction block networks across multiple
	layers of a neural network and trains them to extract representations at
	different scales. 
	The multi-scale aspect of the extraction is included as features of intermediate network layers have not yet been spatially reduced by pooling layers and can appear with different sizes in the feature maps.
	
	Owing to this multi-scale and multi-level nature of extraction, this novel NN
	architecture is called $\mathbf{M^2}$.  Moreover, compelled by the
	representation alignment property of contrastive losses
	\cite{wilfred2007energizing} (i.e., similar samples share similar features), we
	propose maximizing the similarity of same-class representations and
	simultaneously minimizing the similarity between the ones of a different class,
	in each level of extraction. These objectives and the role of the $\mathbf{M^2}$ and contrastive loss components are illustrated in
	Fig. \ref{fig:method_new}.
	
	By adding such a tailored custom contrastive loss term to a cross-entropy loss
	function, the proposed $\mathbf{M^2}$\textbf{-CL} model is able to yield
	superior accuracy in all DG benchmarks and surpass previous state-of-the-art
	models. In the following Sections \ref{background} through
	\ref{network_architecture} we:
	\begin{itemize}
		\item introduce the necessary background and notations for classification in DG,
		\item formulate the intuition behind our proposed framework,
		\item present in detail the mechanisms for extracting multi-scale and multi-level representations 
		\item describe the implementation of the proposed method on different network architectures.
	\end{itemize}
	
	\subsection{Background \& Terminology}
	\label{background}
	Let a Domain $s$ correspond to an (unknown) data distribution of samples in $(\mathcal{X}, \mathcal{Y})$. Domain Generalization algorithms focus on learning a parametric model $f(\cdot; \theta)$, trained on samples $(\pmb{x}^{(s)}, y^{(s)})$ drawn from a set of $N$ \emph{source} Domains $\{S_1, S_2, \dotsc, S_N\}$, that performs well on data $(\pmb{x}^{(t)}, y^{(t)})$ drawn from $K$ \emph{target} Domains $\{T_1, T_2, \dotsc, T_K\}$.
	
	In this paper, we consider domain generalization for the image
	classification task and aim to improve the ability of a model $M_\theta$:
	$\mathcal{X}$ $\rightarrow$ $\mathcal{Y}$ to encode class-specific and domain-specific attributes separately. Then, we want to train an
	Encoder $f_\theta$ that can extract distinguishable and common attributes
	between source and target domains.

	We think of an object, as a mixture of class-relevant and domain-specific attributes (Fig. \ref{fig:part_based}).
	Our main hypothesis is that it is challenging to avoid \emph{entangled}
	representations, i.e. representations containing both invariant information
	about the class and the domain-specific (but class-irrelevant) attributes, when
	relying only on the last layers of a deep CNN. We therefore propose to build
	representations that use features extracted from multiple layers of the network
	and argue that images with common labels should contain and carry similar
	representations across domains. Furthermore, we argue that including several scales of extracted information could assist in capturing features that have not yet been spatially reduced and push the model towards disregarding representations containing non-causal attributes of the image. 
	Moreover, we aim to push the model towards maximizing the similarity of representations extracted from images with the same label while at the same time minimizing the similarity of representations originating from images of different classes. The logic behind the implementation of each component in
	the proposed method is illustrated in Fig. \ref{fig:method_new}.

	\subsection{Extraction block}
	\label{sec:extraction_block}
	
\begin{figure}
		\centering
		\includegraphics[width=\linewidth]{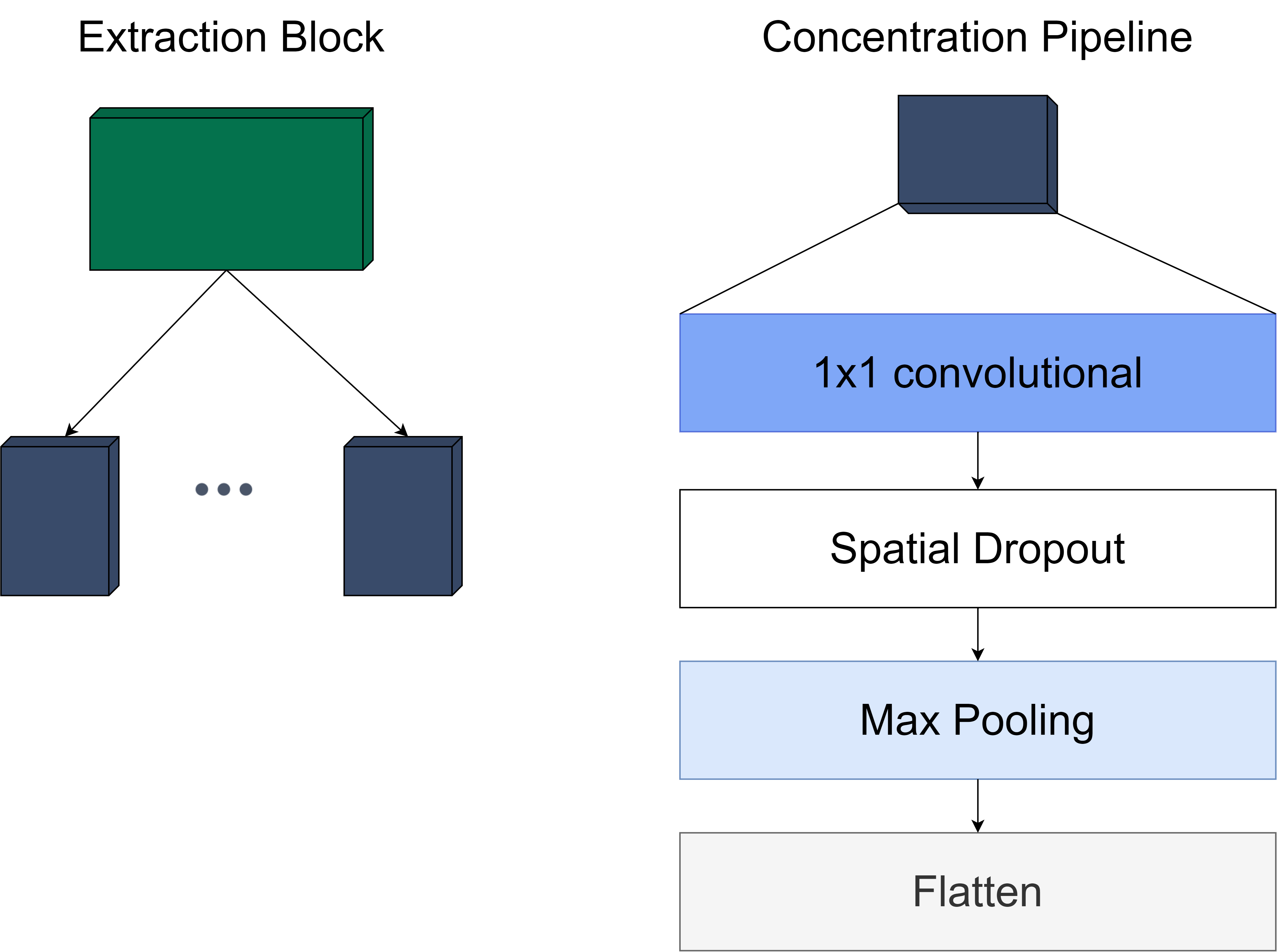}
		\caption{Visualization of the \emph{Extraction Block} and \emph{Concentration Pipeline} implementation. Each block can be connected to
			any intermediate layer of the backbone model, followed by multiple concentration pipelines.
			Each pipeline consists of a ${1 \times 1}$ Convolutional, Spatial Dropout, Max Pooling and Flatten layer, which is thereafter passed through a multilayer perceptron (MLP) and connected to the framework's concatenation layer as depicted in Fig. \ref{fig:model}.}
		\label{fig:extraction_block}
	\end{figure}

	The main mechanism for implementing the feature extraction in $\mathbf{M^2}$ is a custom \textit{extraction block} which can be attached to multiple layers of the network, as illustrated in Fig. \ref{fig:model}. The extraction block consists of a parallel set of \textit{concentration pipelines}, presented in Fig. \ref{fig:extraction_block}, which aim at the multi-scale processing of the intermediate convolutional layer outputs.
	
	More specifically, let $c$ be the number of output channels of a convolutional
	layer. We first apply a $1\times 1$ convolutional layer to reduce the number of
	channels, leading to a more compressed representation. This is controlled by a
	reduction parameter $r$, such that the output channels are $\left\lfloor
	c/r\right\rfloor$.
	
	The convolutional layer is followed by a spatial dropout layer.  This layer has
	been proposed by Tompson et al. \cite{Tompson_2015_CVPR} for the human pose
	estimation problem. In the standard dropout layer, individual features (i.e.,
	pixels) of the output are dropped, but due to the spatial correlation of
	natural images, non-dropped neighboring features carry significant mutual
	information, rendering the dropout ineffective. The spatial dropout layer drops
	entire feature maps, thus avoiding this problem.  Depending on the target class
	and image, a target ``attribute'' may appear at different scales. We therefore
	apply several $k \times k$ Max Pooling layers with stride equal to 1. The pool
	size $k$ depends on the input and is selected such as the output feature maps
	have sizes equal to $8\times 8$, $4\times 4$ and $2 \times 2$, for the earlier
	layers, followed by $7\times 7$ and $3\times 3$ for the layers further down the
	network. We have experimented with both cascading and parallel implementations
	of the above pipeline and have found the parallel architecture to be more
	effective, as described in Section \ref{sec:ablation}. In order to allow our
	model to learn a mapping between the extracted features, we pass each Max
	Pooling layer output through a multilayer perceptron (MLP), which in turn
	yields a vector.
	
	This framework is straightforward to implement and highly adaptable, in the
	sense that the extraction blocks can be connected to any layer of a CNN and one
	can use an arbitrary number of pipelines in each block. Moreover the proposed framework can be leveraged by using different CNN as a backbone.
	
	
	\subsection{Network Architecture}
	\label{network_architecture}
	The proposed architecture of $\mathbf{M^2}$ is illustrated in Fig. \ref{fig:model} and is built on the ResNet-18 architecture. For our method, we select outputs from
	a total of 13 intermediate conv layers of a CNN and pass them through an extraction block, as described in Section \ref{sec:extraction_block}.

	The selected intermediate layers include all residual half block layers (i.e.,
	those leading to reduction of the output size), as well as selected
	convolutional layers. All later layers of the network are included. The output of each concentration pipeline is concatenated into a single feature vector which passes through a classification head with a single fully-connected layer.
	
	\begin{figure*}[t!]
		\centering
		\includegraphics[width=\linewidth]{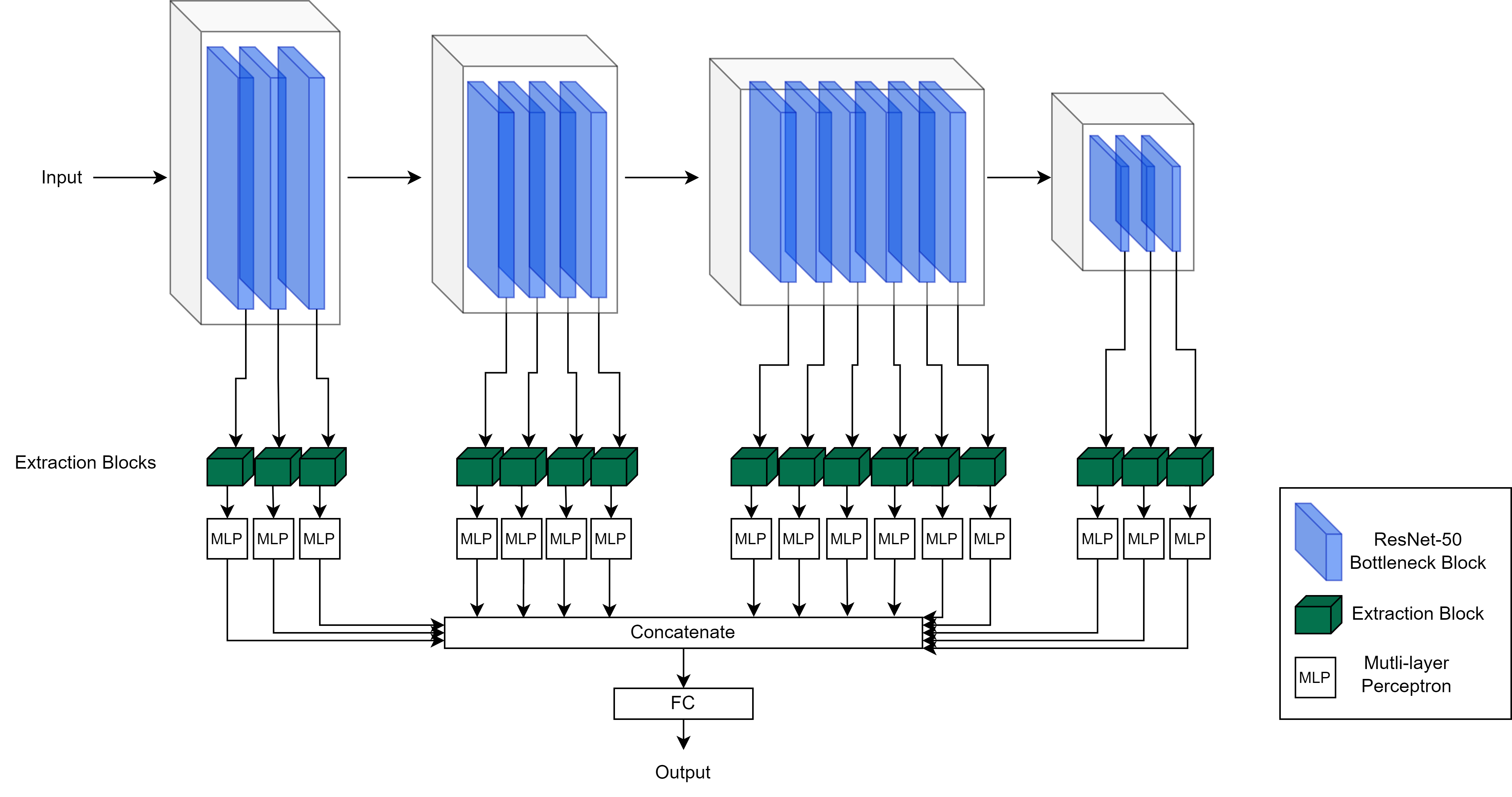}
		\caption{Visualization of the M\textsuperscript{2} architecture built on a ResNet-50 model. Similar to the ResNet-18 implementation in Figure \ref{fig:model}, we once again extract feature maps by using multiple \textit{extraction blocks} (green boxes). Our proposed extraction blocks are connected to the last convolutional layer of each of the ResNet's bottleneck blocks.}
		\label{fig:res50_model}
	\end{figure*}
	
	To further demonstrate the effectiveness of our method, we also implement our algorithm with a larger ResNet-50 model, in which the ratio of the additional parameters of the introduced extraction block and the backbone model is smaller, when compared to the ResNet-18 implementation. In a more straightforward approach, each extraction block is connected to the output of the last convolutional layer in each one of the model's bottleneck blocks, as shown in Figure \ref{fig:res50_model}.
	
	\subsection{A Contrastive Loss for Multi-Level Representations}
	\label{contrastive_loss}
	
	Consider the representation $\pmb{u}^{(l)}_i$ of sample $\pmb{x}_i$, which is the extracted and processed (via an extraction block) output of layer $l$ of the neural network after normalization, such that $\lVert
	\pmb{u}^{(l)}_i\rVert = 1$. We define the following probability measure of class
	$c$ for a batch of $N_b$ elements under this representation as
	\begin{equation}
		p^{(l)}(c) = \frac{\sum\limits_{i =  1}^{N_c}\sum\limits_{\substack{j=i+1\\y_i=y_j=c}}^{N_c}e^{\pmb{u}^{(l)T}_{i}\pmb{u}^{(l)}_j / \tau}}
		{\sum_{k=1}^{N_b} \sum_{m=l+1}^{N_b} e^{\pmb{u}^{(l)T}_k\pmb{u}^{(l)}_m / \tau}}
		\label{eq:class_likelihood}
	\end{equation}
	where $N_c$ is the total number of elements of class $c$ in the batch and
	$\tau$ is a temperature hyperparameter, similar to the terms in \cite{chen2020simple, he2020momentum, grill2020bootstrap}. $\tau$ is an important hyperparameter in both contrastive \cite{wu2018unsupervised} and fully supervised learning \cite{hinton2015distilling}, as it controls the density of the probability mass distribution. Note that
	Eq. \eqref{eq:class_likelihood} increases if images of the same class have
	similar representations and images of different classes have dissimilar
	representations.
	
	Considering all $C$ classes, the probability becomes
	\begin{equation}
		\mathcal{P}^{(l)} = \prod_{c=1}^C p^{(l)}(c)
		\label{eq:layer_likelihood}
	\end{equation}
	
	We can therefore take the logarithm of $\mathcal{P^{(l)}}$ to introduce the following
	loss for layer $l$:
	\begin{equation}
		\mathcal{L}^{(l)} = -\log\mathcal{P}^{(l)} =
		-\sum_{c=1}^{C}\log p^{(l)}(c)
		\label{eq:layer_loglikelihood}
	\end{equation}
	The learning objective is therefore to maximize the probability of all classes under this representation, or minimize $\mathcal{L}^{(l)}$, while at the same time maintaining the model's effectiveness in discriminating the target
	classes. This leads to the following loss function
	\begin{equation}
		\mathcal{L} = \mathcal{L}_{CE} + \alpha\sum\limits_{l=1}^{\mathcal{M}}\mathcal{L}^{(l)}
		\label{eq:loss}
	\end{equation}
	where $\mathcal{M}$ is the total number of layers which are considered for the
	representation, $\mathcal{L}_{CE}$ is the cross-entropy loss and $\alpha$ is
	a hyperparameter indicating the magnitude of importance the additional objective holds in model training.
	
	Computationally, note that if $\pmb{U}^{(l)}$ is a matrix containing the vectors
	$\pmb{u}^{(l)}$ of the batch, one per row, then the inner products
	$\pmb{u}^{(l)T}_i\pmb{u}^{(l)}_j$ of Eq. \eqref{eq:class_likelihood} are the
	elements of matrix $\pmb{U}^{(l)T}\pmb{U}^{(l)}$. Moreover, the denominator is the same for all classes. Therefore, the inner products of the denominator of the proposed loss \eqref{eq:loss} only needs to be computed once for each batch.
	
	
	
	\section{Experiments}
	In our experiments, we opted for vanilla ResNet-18 and ResNet-50 \cite{He_2016_CVPR} models, pre-trained on ImageNet, as the backbone networks of our method. We implemented our framework with the PyTorch library \cite{NEURIPS2019_9015}, on one NVIDIA RTX A5000 GPU. For the optimizer, we used SGD and trained for 30 epochs on every dataset and set the learning rate to 0.001. For our method we set the reduction parameter $r$ equal to 4 and select to implement 3 concentration pipelines per block in the early layers and 2 pipelines per block in the later ones for both the ResNet-18 and ResNet-50 implementations. Depending on the input of each Max Pooling layer in the concentration pipelines, we set the pool size \emph{k} such that the early layers return feature maps of size $8\times 8$, $4\times 4$ and $2\times 2$. Due to the size of the later convolutional layers (i.e., in the last residual block), we configure the max pooling layers to produce feature maps of size $7\times 7$ and $3\times 3$. In all experiments containing our custom loss we set the $\alpha$ hyperparameter 0.01 and the temperature $\tau$ to 1.0, as default values. The value of $\alpha$ reflects the common practice in previous works proposing regularization methods \cite{loshchilov2017decoupled}. Moreover, as different values of $\tau$ could affect the performance of our model in each dataset, we chose to set it to 1.0 for a fair comparison against all datasets and explore its importance in separate experiments. To this end, Section \ref{sec:ablation} presents a sensitivity analysis for hyperparameters $\alpha$ and $\tau$. Finally, for all implementations we use a standard batch size of 128 images.
	
	\subsection{Datasets}
	To evaluate our method, we run experiments on 4 publicly available datasets,
	PACS \cite{Li_2017_ICCV}, VLCS \cite{5995347}, Office-Home \cite{venkateswara2017deep} and NICO \cite{HE2021107383}.
	
	\textbf{PACS} contains 7 categories of images from 4 different domains
	(\textbf{P}hoto, \textbf{A}rt Painting, \textbf{C}artoon and
	\textbf{S}ketch). The total amount of images is 9,991.
	
	\textbf{VLCS} combines 10,729 real-world images of 5 classes from the PASCAL
	\textbf{V}OC, \textbf{L}abelMe, \textbf{C}altech 101 and \textbf{S}UN09
	datasets.
	
	
	
	\textbf{Office-Home}, like PACS, is composed of images from the art, clipart,
	product and and real-world domains. The dataset contains 15,588 examples and 65
	classes.
	
	\textbf{NICO} was recently proposed as non-iid image classification dataset
	capable of evaluating algorithms on out-of-distribution settings. It consists
	of a total of 25,000 images from 19 classes, 10 of which are types of animals
	and 9 types of vehicle. NICO simulates real-world conditions by separating each
	class of images into contexts (e.g dog on beach, horse running, boat with
	people, etc.).
	
	For PACS, VLCS and Office-Home we follow the standard
	\emph{leave-one-domain-out cross-validation} protocol, as described in \cite{Li_2017_ICCV, Ghifary_2015_ICCV} and \cite{gulrajani2021in} , by holding out one domain as the test
	split and training our model on the remaining data. 
	Similarly to the 3 previous datasets, for NICO we evaluate our model by
	randomly selecting 3, 5 and 7 contexts in each class and holding them out as a
	test set, simulating the leave-one-domain-out, or in this case
	leave-mutliple-domains-out setting.

	\begin{table*}\centering
		\begin{center}
			\caption{Top-1\% accuracy results on the \textbf{PACS} (left) and \textbf{VLCS} (right) datasets. The columns denote the target domains. The top results are highlighted in \textbf{bold} while the second best are \underline{underlined}.}
			\label{tab:pacs_vlcs}
			\begin{tabular}{c|cccc|c||cccc|c}
				\toprule
				\noalign{}
			\textbf{Method} & \textbf{Art} & \textbf{Cartoon} & \textbf{Photo} & \textbf{Sketch} & \textbf{Avg} 
			& \textbf{Caltech} & \textbf{Labelme} & \textbf{Sun} & \textbf{Voc} & \textbf{Avg}\\
			\midrule
			\textit{ResNet-18} &&&&&&&&&&\\
			ERM    
			& 79.68 & 77.76 & 88.91 & 75.72 & 80.53 
			& 84.80 & 59.05 & 65.27 & 70.12 & 69.81 \\
			RSC    
			& 78.21 & 74.30 & 92.51 & \underline{78.75} & 80.94
			& 88.69 & \underline{62.91} & 68.24 & 68.95 & 72.34 \\
			CORAL  
			& 76.03 & 75.60 & 93.23 & \textbf{78.78} & 80.91
			& 95.67 & 60.32 & 69.95 & 72.82 & 74.69 \\
			MIXUP   
			& 79.19 & 74.09 & 95.36 & 73.95 & 80.65 
			& 96.20 & 61.77 & 69.76 & 72.93 & 75.17 \\
			MMD    
			& 75.71 & 66.47 & 94.08 & 69.59 & 76.46
			& 95.67 & 58.58 & 60.70 & 68.04 & 70.75 \\
			SagNet 
			& 77.24 & 77.10 & 93.42 & 73.28 & 80.26
			& 89.39 & 61.11 & 68.20 & 73.86 & 73.14 \\
			SelfReg 
			& 79.68 & \underline{77.91} & 94.61 & 74.64 & 81.71
			& 96.11 & 60.51 & 68.12 & 74.51 & 74.81 \\
			ARM   
			& 79.21 & 75.74 & 92.82 & 75.09 & 80.71
			& 96.64 & 59.88 & 68.10 & 71.31 & 73.98 \\
			EQRM 
			& \underline{81.39} & 74.57 & 93.41 & 75.60 & 81.24
			& 96.11 & 61.12 & 66.56 & 71.40 & 73.80 \\
			SAGM 
			& 79.21 & 77.35 & 94.61 & 79.58 & 82.68
			& 96.81 & 60.07 & 69.15 & 73.04 & 74.77 \\
			\midrule
			
			$M^2$ \textsubscript{(Ours)}
			& 79.86 & \underline{77.91} & \underline{95.65} & 77.90 & \underline{82.83}
			& \underline{97.61} & 60.18 & \underline{70.75} & \underline{75.89} & \underline{76.11} \\
			$M^2$-CL \textsubscript{(Ours W/Loss)}
			& \textbf{81.66}  & \textbf{78.42} & \textbf{97.00} & 77.07 & \textbf{83.54}
			& \textbf{98.23}  & \textbf{63.85} & \textbf{72.35} & \textbf{76.67} & \textbf{77.78}\\
			
			\midrule
			
			\textit{ResNet-50} &&&&&&&&&&\\
			ERM    
			& 83.90 & 78.60 & 97.30 & 73.50 & 83.33 
			& 97.80 & 63.30 & 70.30 & 75.90 & 76.83 \\
			RSC    
			& 81.66 & \underline{80.86}	& 96.18	& 75.79	& 83.62
			& 95.40	& 64.65	& 70.45	& 73.33	& 75.96 \\
			CORAL  
			& 86.00	& 75.50	& 96.20	& 76.60	& 83.58
			& 97.50	& 64.00	& 69.70	& 76.70	& 76.98\\
			MIXUP   
			& \textbf{88.00} & 74.30 & 97.20	& 75.30	& 83.70 
			& \underline{97.90}	& \textbf{65.50}	& \textbf{73.30} & \textbf{77.80} & \textbf{78.62} \\
			MMD    
			& 85.90	& 78.10 & 96.20	& 71.10	& 82.82
			& 97.70	& 63.10	& 68.60	& 77.40	& 76.72\\
			SagNet 
			& 85.47 & 80.09	& 94.46	& \underline{77.83} & 84.46
			& 97.17 & 64.00	& 70.50	& 73.48	& 76.29\\
			SelfReg 
			& 83.61 & 79.15 & 95.80 & \textbf{78.10} & 84.16
			& 94.69 & 64.47 & 68.88 & 73.82 & 75.46\\
			ARM   
			& 83.22 & 80.01	& 95.13	& 76.75	& 83.78
			& 96.55	& \underline{65.16}	& 70.18	& 73.63	& 76.38\\
			EQRM 
			& 86.82 & 79.85 & 94.91 & 78.09 & 84.91 
			& 97.17 & 63.38 & 69.53 & 74.37 & 76.11 \\
			SAGM 
			& 85.33 & 80.55 & 95.88 & 80.06 & \underline{85.45} 
			& 97.87 & 64.97 & 70.57 & 76.29 & 77.42 \\
			\midrule
			$M^2$ \textsubscript{(Ours)}
			& 86.29	& 77.56	& \underline{97.79} & 75.41 & 84.26
			& 97.79	& 64.33	& 70.33	& 74.78	& 76.81 \\
			$M^2$-CL \textsubscript{(Ours W/Loss)}
			& \underline{87.25}	& \textbf{81.84} & \textbf{98.50} & 76.30 & \textbf{85.97}
			& \textbf{98.23}	& \textbf{65.50} & \underline{72.25}	& \underline{77.44}	& \underline{78.36} \\
			\hline
			
		\end{tabular}
	\end{center}
\end{table*}

\begin{table*}\centering
	\begin{center}
		\caption{Top-1\% accuracy results on the \textbf{Office-Home} (left)
			and \textbf{NICO} (right) datasets. For Office-Home, the columns
			denote the target domains. For NICO, the number of left out contexts is
			denoted by $N$. The top results are highlighted in \textbf{bold}
			while the second best are \underline{underlined}.}
		\label{tab:office_nico}
		\begin{tabular}{c|cccc|c||ccc}
			\midrule\noalign{}
			\textbf{Method} 
			& \textbf{Art} & \textbf{Clipart} & \textbf{Product} & \textbf{Real-World} & \textbf{Avg}
			& \textbf{N=3} & \textbf{N=5} & \textbf{N=7}\\
			\noalign{}
			\midrule
			\noalign{}
			\textit{ResNet-18} &&&&&&&&\\
			ERM     
			& 47.88  & 45.64 & 67.34 &  67.52 & 57.10
			& 83.77  & 79.53 & 74.53 \\
			RSC     
			& 54.43  & 46.89 & 67.84 & 70.53  & 59.92
			& 81.36  & 77.30 & 74.52 \\
			CORAL  
			& 53.86 & 46.96 & 68.88 & 71.87 & 60.39 
			& 84.15 & 79.86 & 76.31 \\
			MIXUP   
			& 53.35 & 48.13 & 68.35 & 70.51 &  60.08
			& 84.01 & 80.20 & 78.90 \\
			MMD
			& 48.86 & 44.67 & 66.83 & 66.37 & 56.68    
			& 84.33 & 79.23 & 77.38 \\
			SagNet
			& 52.52 & 47.99 & 67.30 & 70.83 & 59.66
			& 85.09 &  82.23  & 79.05 \\
			SelfReg 
			& 54.22 & 48.11 & 68.43 & 71.83 & 60.65
			& \underline{86.27} & \underline{83.23} & 81.40\\
			ARM
			& 49.48 & 45.19 & 63.73 & 68.36 & 56.69
			& 84.54 & 80.76 & 77.43 \\
			EQRM 
			& 55.67 & 47.85 & 69.76 & 71.29 & 61.14 
			& 84.93 & 80.81 & 76.32 \\
			SAGM 
			& 53.14 & \underline{48.22} & \underline{70.74} & 71.25 & 60.83
			& 85.02 & 81.94 & 79.33 \\
			\midrule
			$M^2$ \textsubscript{(Ours)}
			& \underline{56.91} & 47.77 & 68.33 & \underline{72.21}   & \underline{61.29}
			& 86.02    & 82.81 & \underline{81.64} \\
			$M^2$-CL \textsubscript{(Ours W/Loss)} 
			& \textbf{58.55}  & \textbf{49.57} & \textbf{71.53} & \textbf{73.43} & \textbf{63.27}
			&  \textbf{87.93}    & \textbf{84.10}    & \textbf{82.14} \\ 
			
			\midrule
			
			\textit{ResNet-50} &&&&&&&&\\
			ERM     
			& 62.30 & 54.10 & 75.30 & 77.40 & 67.27
			& 87.02	& 83.66 & 82.04 \\
			RSC     
			& 61.30 & 52.50 & 74.38 & 76.24  & 66.10
			& 87.78	& 83.40	 & 80.01 \\
			CORAL  
			& 64.40 & 55.40	& 76.20	& 78.40	& 68.60 
			& 87.15 & 84.91	& 83.64 \\
			MIXUP   
			& 63.80	& 52.90	& 77.30	& 78.70	& 68.18
			& 88.15 & 83.28 & 82.03 \\
			MMD
			& 62.20	& 52.70	& 75.50	& 78.10	& 67.13    
			& 88.38	& 83.69	& 83.04 \\
			SagNet
			& 63.50	& 52.89	& 74.07	& 75.21	& 66.42
			& 85.96	& 83.62	& 81.72 \\
			SelfReg 
			& 60.60 & 53.32 & 72.07 & 76.92 & 65.73
			& 86.78 & 85.67 & 84.59 \\
			ARM
			& 56.18	& 52.23	& 72.21	& 73.36	& 63.49
			& 87.32	& 82.67 & 82.38 \\
			EQRM 
			& 65.15 & 55.49 & 76.89 & 78.25 & 68.94 
			& \textbf{89.55} & 85.92 & 83.07 \\
			SAGM 
			& 63.29 & 54.26 & 76.37 & 77.84 & 67.94 
			& 87.79 & 85.27 & 84.32 \\
			\midrule
			$M^2$ \textsubscript{(Ours)}
			& \underline{65.36} & \underline{55.55} & \underline{78.15} & \underline{79.50}   & \underline{69.64}
			& 88.78 & \underline{86.88} & \underline{85.84} \\
			$M^2$-CL \textsubscript{(Ours W/Loss)}
			& \textbf{68.45}  & \textbf{57.04} & \textbf{78.66} & \textbf{80.14} & \textbf{71.07}
			&  \underline{89.30}    & \textbf{87.68}    & \textbf{86.90} \\ 
			
			\hline
		\end{tabular}
	\end{center}
\end{table*}

\subsection{Baselines}
For a baseline we compare our framework with 10 other previous state-of-the-art
methods, which utilize a ResNet-18 and ResNet-50 as their backbone model.
To demonstrate the effectiveness of our model we chose to evaluate against both multi-source and single-source DG methods. The compared methods are ERM \cite{vapnik1999nature}, RSC \cite{huangRSC2020}, MIXUP \cite{yan2020improve}, CORAL\cite{sun2016deep}, MMD \cite{li2018domain}, SagNet \cite{nam2021reducing}, SelfReg \cite{kim_selfreg_2021}, ARM \cite{zhang2021adaptive}, EQRM \cite{eastwood2022probable} and finally SAGM \cite{wang2023sharpness}. It is 
important to mention, that even though ensemble learning methods such as SIMPLE \cite{li2023simple}, MIRO \cite{avidan_domain_2022}, and PCL \cite{yao_pcl_2022} provide noteworthy improvement in performance, they are not evaluated against our methods, for fair comparison. However, the proposed 
models could be included in ensemble learning frameworks for improved results.
All the baselines are implemented and executed using the
codebase of DomainBed \cite{gulrajani2021in}, while the training, validation and test data splits remain the same across all experiments. The hyperparameters of each method is set as recommended by the authors of their respective papers. In our results, we present the average accuracy over 3 runs.

\subsection{Results}

Table \ref{tab:pacs_vlcs} summarizes the results of our experiments on PACS and
VLCS, while Table \ref{tab:office_nico} on Office-Home and NICO. When compared
to previously proposed DG methods, $\mathbf{M^2}$\textbf{-CL} is able to set the state-of-the-art in every dataset and thus prove some degree of robustness in its learning and inference processes.

In \textbf{PACS}, our model is able to surpass the previously proposed methods
with both the ResNet-18 and ResNet-50 base models. On average, our algorithm exceeds the second best performing models by 1.83\% and about 0.5\% respectively. With the exception of the \textit{Sketch} domain in the ResNet-18 case, the addition of the custom loss term boosts our model's performance in all other domains. We believe that since the white background is dominant in the Sketch domain, our model is not able to disentangle the important attributes in the images.

In the \textbf{VLCS} dataset, we find that our method again sets the state-of-the-art in all domains when adopting the ResNet-18 base model and is highly competitive in the ResNet-50 implementation, surpassing all predecessors in two out of four domains and achieving the second-best performance in the remaining two. On average, our ResNet-18 model is able to achieve an increase of 2.61\% in accuracy, while the ResNet-50 model's performance is comparable to the top score. Once again, our custom loss increases the model's accuracy in both cases.

The same trend follows in \textbf{Office-Home} where our model surpasses the
baselines by 3.61\% and 2.47\%, for the ResNet-18 and ResNet-50 implementations respectively. Additionally, both of our models set the state-of-the-art in all of the domains. 

As far as \textbf{NICO} is concerned, our method continues to demonstrate an improvement in 2 out 3 settings over the baselines. With the exception of the case when 3 contexts are left out during the ResNet-50 training, our model consistently surpasses the previous algorithms in every setting. When leaving out 3, 5, and 7 contexts, the best ResNet-18 models surpass the other algorithms by a total of 1.66\%, 0.87\% and 3.09\% respectively, while the ResNet-50 models achieve the top scores when leaving out 5 (+1.76\%) and 7 (+2.58\%) contexts.

\begin{figure}[t]
	\centering
	\includegraphics[width=0.78\columnwidth]{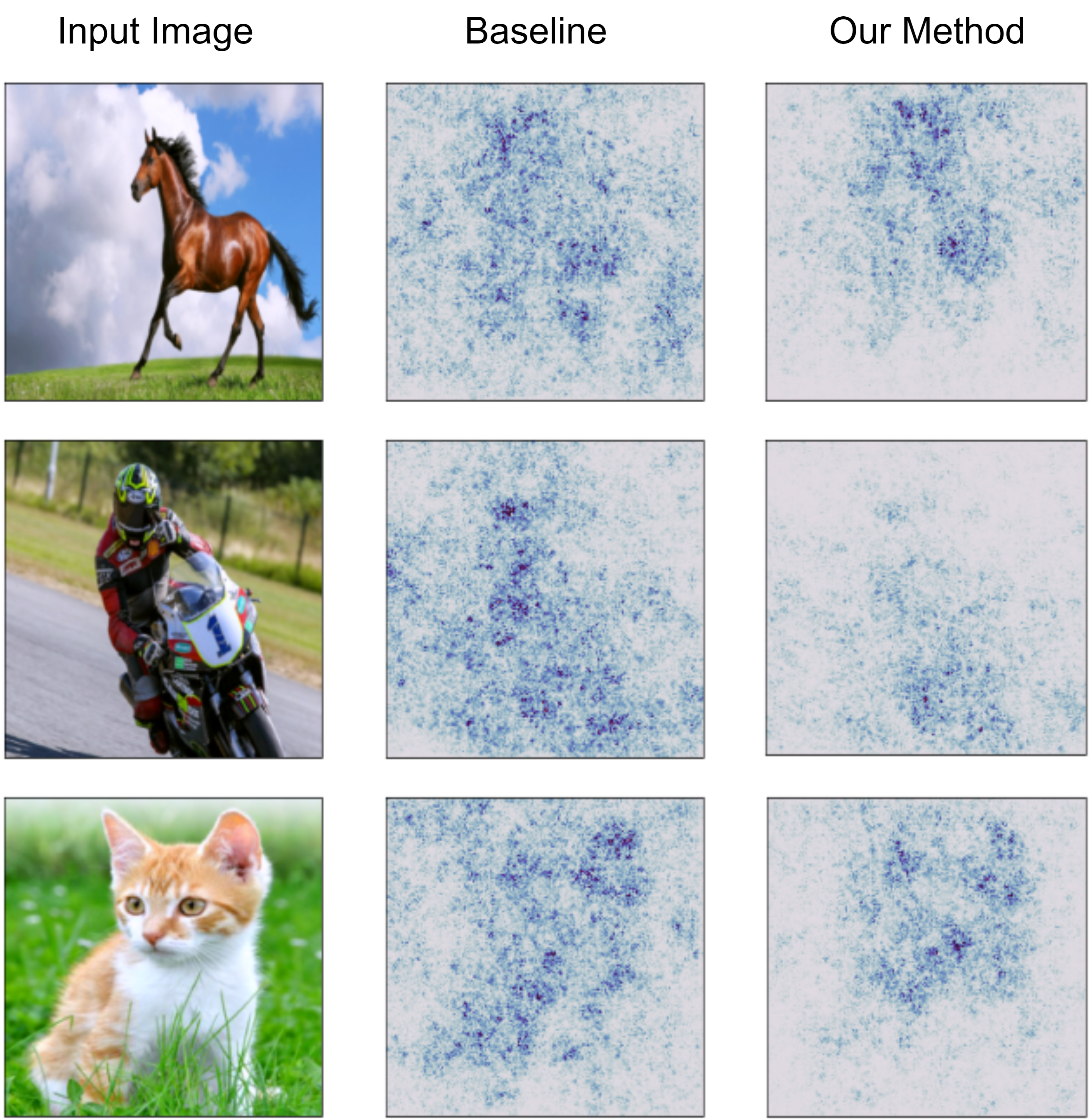}
	\caption{Visualization of saliency maps, produced by the baseline ERM
		ResNet-18 model and our method from images in the NICO dataset. The magnitude of darkness in the saliency map defines the importance of each corresponding pixel to the predicted class, i.e. the darker the pixel, the more it contributes to the final prediction. The baseline ERM model tends to infer based on the background and spurious correlations of the object. On the contrary, our method focuses on the invariant features and ignores the entangled parts (e.g rider of the
		motorcycle).}
	\label{fig:saliency}
\end{figure}

\begin{figure}[t]
	\centering
	\includegraphics[width=0.75\columnwidth]{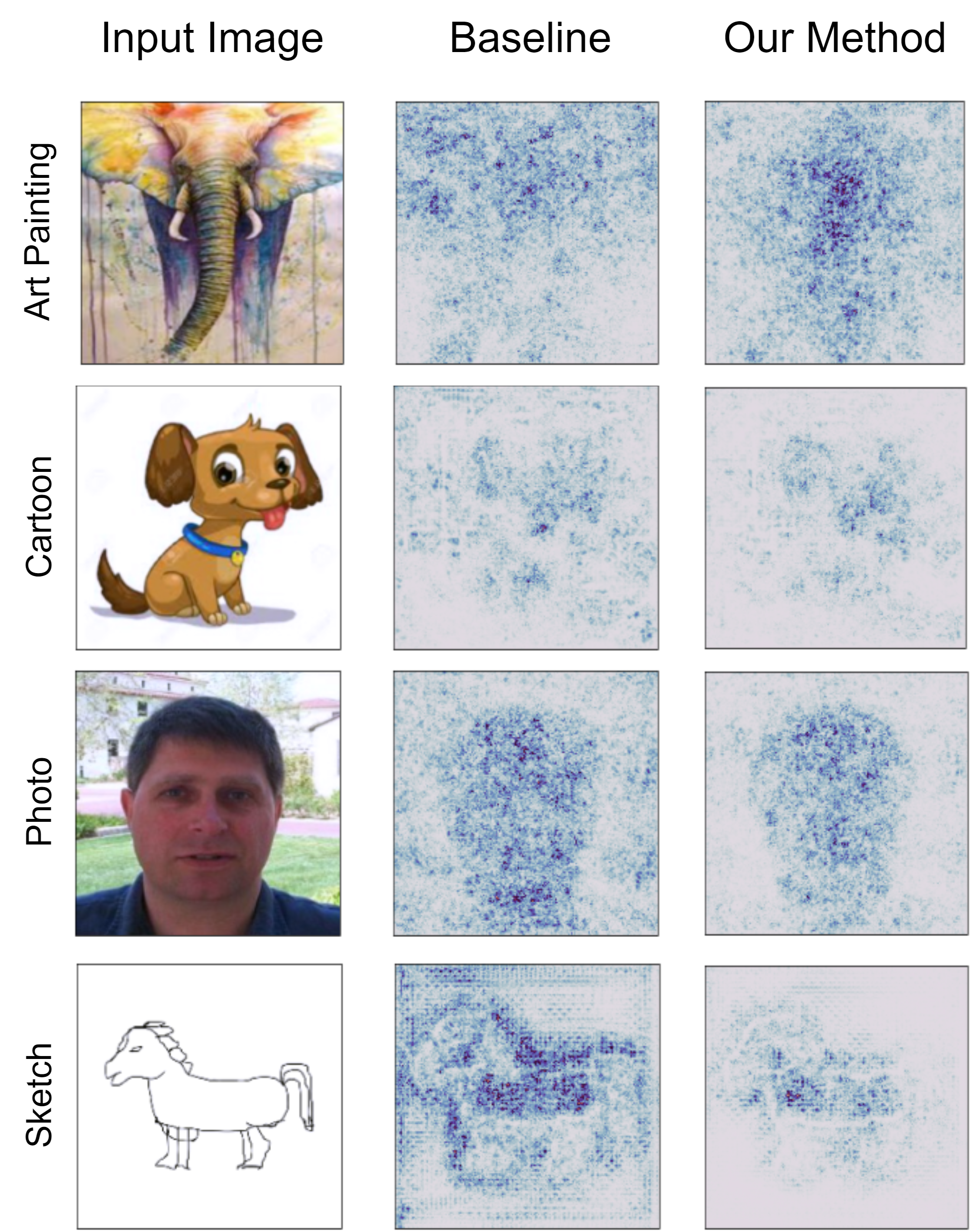}
	\caption{Visualization of saliency maps, produced by the baseline ERM ResNet-18 model and our method, from images in the PACS dataset. Our approach successfully disregards the white backgrounds in the \emph{Cartoon} and \emph{Sketch} domains, and spurious correlations in
		the \emph{Art Painting} and \emph{Photo} domains.}
	\label{fig:saliency_pacs}
\end{figure}

\subsection{Ablation Study \& Sensitivity Analysis}
\label{sec:ablation}

To justify the selection and effectiveness of the components in our framework,
we conduct an extensive analysis and ablation study. Our implementation has 3
key components: a) the compression method in the extraction block, b) the
reduction parameter `$r$' and c) the Spatial Dropout layer. We also take
into consideration d) our proposed loss function \eqref{eq:loss}. Table \ref{ablation_block} summarizes the results of our analysis on the PACS and VLCS datasets. The top set of parameters were also used for the experiments in Office-Home and NICO.

\begin{table*}\centering
	\begin{center}
		\caption{Ablation study on key components of our method, with a ResNet-18 base model, on the PACS and VLCS datasets. We denote the method of reduction in the
			extraction block as {\it pipe} ({\it c} for cascading and {\it p} for parallel), the arbitrary reduction parameter as
			$r$ and the Spatial Dropout layer as {\it drop}. A, C, P and S correspond to the Art Painting, Cartoon, Photo and Sketch domains in PACS, as C, L, V and S, denote the CALTECH 101, LABELME, SUN09 and PASCAL VOC datasets in VLCS.}
		\label{ablation_block}
		\begin{tabular}{cccc||cccc|c||cccc|c}
			\hline\noalign{\smallskip}
			pipe & $r$ & drop & loss
			& A & C & P & S & Avg
			& C & L & S & V & Avg \\
			\noalign{\smallskip}
			\hline
			\noalign{\smallskip}
			c & 2 &     -   & -
			& 75.98 & 73.68 & 94.39 & 71.27 & 78.83 
			& 96.07 & 61.26 & 69.05 & 65.38 & 72.94 \\
			c & 4 &     -   & -
			& 77.50 & 73.96 & 94.03 & 70.23 & 78.93    
			& 95.59 & 59.47 & 70.55 & 66.34 & 72.99 \\
			c & 6 &     -   & -
			& 77.37 & 74.09 & 94.71 & 73.43 & 79.90
			& 95.91 & 58.55 & 69.35 & 66.38 & 72.55 \\
			c & 2 & \checkmark & -
			& 76.30 & 73.94 & 94.75 & 76.05 & 80.26
			& 97.53 & 58.87 & 68.77 & 66.17 & 72.84 \\
			c & 4 & \checkmark & -
			& 77.86 & 74.96 & 94.24 & 76.51 & 80.89
			& 96.08 & 59.25 & 70.47 & 67.22 & 66.50 \\
			c & 6 & \checkmark & -
			& 78.13 & 74.41 & 94.24 & 75.06 & 80.46
			& 96.08 & 60.19 & 69.85 & 67.91 & 73.51 \\
			p & 2 &     -   & -  
			& 78.70 & 59.59 & 69.52 & 64.84 & 68.16 
			& 96.57 & 59.51 & 69.03 & 74.62 & 74.93 \\
			p & 4 &     -   & -  
			& 76.93 & 74.75 & 94.42 & 73.54 & 79.92 
			& 95.95 & 60.16 & 69.55 & 65.89 & 72.89\\
			p & 6 &     -   & -
			& 76.45 & 73.63 & 94.19 & 72.59 & 79.21    
			& 95.43 & \underline{60.22} & 69.73 & 66.10 & 72.87 \\
			p & 2 & \checkmark & -
			& 77.37 & 75.84 & 87.37 & 77.68 & 79.56
			& 97.28 & 59.99 & 70.68 & 65.90 & 73.46 \\
			p & 6 & \checkmark & -
			& 77.37 & 74.83 & 94.83 & 76.56 & 80.89
			& 96.21 & 59.69 & 69.80 & 67.02 & 73.18 \\
			\hline
			p & 4 & \checkmark  & -  
			& \underline{79.86} & \underline{77.91} & \underline{95.65} 
			& \textbf{77.90}    & \underline{82.83}
			& \underline{97.61} & 60.18             & \underline{70.75} 
			& \underline{75.89} & \underline{76.11} \\
			\hline 
			p & 4 & \checkmark  & \checkmark   
			& \textbf{81.66}    & \textbf{78.42} & \textbf{97.00} 
			& \underline{77.07} & \textbf{83.54}
			& \textbf{98.23}    & \textbf{63.85} & \textbf{72.35} 
			& \textbf{76.67}    & \textbf{77.78}\\
			
			\hline
		\end{tabular}
	\end{center}
\end{table*}

\begin{table}\centering
	\begin{center}
		\caption{Sensitivity analysis for hyperparameters $\tau$ and $\alpha$ of the proposed loss function (Eq. \ref{eq:class_likelihood} and \ref{eq:loss}) in $M^2$-CL. The average accuracy of each model across target domains is presented for the PACS and VLCS datasets. The base network for each different setup is a ResNet-18, as in the ablation study of Table \ref{ablation_block}.}
		\label{sensitivity}
		\begin{tabular}{c|c|c||c|c|c}
			\hline\noalign{\smallskip}
			\textbf{$\tau$} &   PACS   &  VLCS & \textbf{$\alpha$} & PACS & VLCS\\
			\noalign{\smallskip}
			\hline
			\noalign{\smallskip}
			0.01   &   81.88  & 77.57 
			&&& \\
			0.1    &   82.20  & 77.48 
			& \multirow{2}{*}{0.00} & \multirow{2}{*}{82.83} & \multirow{2}{*}{76.11} \\
			0.2    &   83.82  & 77.58 &&&\\
			0.4    &   \underline{85.38}  & 75.31 
			& \multirow{2}{*}{10e-5} & \multirow{2}{*}{82.35} & \multirow{2}{*}{76.43}\\
			0.6    &   84.97  & 75.88 
			&&& \\
			0.8    &   85.30  & 77.71 
			& \multirow{2}{*}{10e-4} & \multirow{2}{*}{\underline{83.12}} & \multirow{2}{*}{\underline{77.59}} \\
			1.0    &   83.54  & 77.78 
			&&& \\
			1.2    &   84.75  & 77.25 
			& \multirow{2}{*}{10e-3} & \multirow{2}{*}{\textbf{83.54}} & \multirow{2}{*}{\textbf{77.78}}\\
			1.4    &   85.07  & 77.54 
			&&&\\
			1.6    &   \textbf{85.41}  & \underline{77.80} 
			& \multirow{2}{*}{10e-2} & \multirow{2}{*}{80.52} & \multirow{2}{*}{77.01}\\
			1.8    &   83.67  & \textbf{78.10}
			&&&\\
			2.0    &   82.17  & 77.29 
			& \multirow{2}{*}{10e-1} & \multirow{2}{*}{67.26} & \multirow{2}{*}{56.92}\\
			10.0   &   83.31  & 77.32 &&&\\
			100.0  &   82.04  & 76.69 &&&\\
			
			
			\hline
		\end{tabular}
	\end{center}
\end{table}


The compression in the extraction block can be implemented either via {\it
	cascading} or {\it parallel} concentration pipelines. When adopting the
cascading method, each max pooling layer is connected to the same $1\times1$
convolutional layer in the extraction block, thus creating a single
pipeline. On the contrary, in the parallel implementation, each pooling layer
is connected to a separate conv layer, creating several parallel
pipelines. Although the cascading pipelines have a slight advantage in specific
domains, it is apparent that the parallel pipelines perform significantly
better overall.

The reduction parameter `$r$', controls the reduction ratio of feature
maps that will be extracted from each intermediate layer in the backbone
network. For example, if the intermediate layer of the backbone network
produces 128 feature maps and we set `$r$' equal to 4, the feature maps passed
to each concentration pipeline will be 32. As $r$ is an arbitrary number, we
select to experiment for values of 2, 4 and 6. As shown in Table 4, setting
$r=4$ consistently improves the overall performance of the model in both the
PACS and VLCS datasets.

The third key component in our framework is the {\it Spatial Dropout}
layer. We added this regularization layer in order to account for the
correlation between adjacent pixels in the reduced feature maps. Our analysis
shows that the dropout layer indeed assists the generalization ability of our
methodology, as it boosts its performance by 1-1.8\% on average in PACS and
1-2.5\% in VLCS. Moreover, it is apparent that the addition of our proposed custom loss to the model proves highly effective, as it is able to increase its performance by a clear margin in every domain except Sketch. 

The proposed $\mathbf{M^2}$\textbf{-CL} model has two hyperparameters of its own: $\tau$ and $\alpha$. In order to research the behavior of our method and its sensitivity to changes in the above parameters, we performed additional experiments on the PACS and VLCS datasets. As a first step, we held $\alpha$ to the default value of 0.01 and set $\tau$ in $[0.1, 2.0]$ while also experimenting with extreme values, such as $0.01$, $10$ and $100$. After analyzing the behavior of $\mathbf{M^2}$\textbf{-CL} regarding $\tau$, we followed by setting $\alpha$ in $[10e-5, 10e-1]$ and keeping $\tau$ to its own default value of $1.0$. The results of the analysis are presented in Table \ref{sensitivity}, which reports the average accuracy of each separate setup. The variation in the values of $\tau$ clearly affect the $\mathbf{M^2}$\textbf{-CL} model, as its performance fluctuates between $-1.66\%$ and $+1.87\%$ in PACS, when compared to the presented results of Table \ref{tab:pacs_vlcs}. In VLCS however, the accuracy of $M^2$-CL is less sensitive to changes in $\tau$, as the results do not diverge significantly from those of the default model. With regard to $a$, the results of the analysis were somewhat expected. When setting $\alpha$ to a value close to zero, such as $10e-05$ , the models' accuracy falls back to the performance of $M^2$ which does not utilize the proposed custom loss. Contrarily, when $\alpha=1.0$ both learning objectives in Eq. \ref{eq:loss} are of equal strength, which is not optimal for training $\mathbf{M^2}$\textbf{-CL}. As a result, its performance greatly degrades, yielding even up to $-20\%$ accuracy.

\subsection{Qualitative Results}
To validate our assumptions and intuition regarding the representation extraction capabilities of our model, we choose to visualize the gradients
of the class score function with respect to the input pixels of a given image,
as proposed in \cite{simonyan2014deep}. By using saliency maps, one can
visualize the pixels that affect the classification the most. The resulting
figures illustrate that the baseline model's predictions are strongly
influenced by the various contexts and spurious correlations in the input
images, while our method focuses on the invariant features. We select to
produce saliency maps on images that were not in the training split in an
attempt to assess the robustness of our approach.

Fig \ref{fig:saliency} shows saliency maps of images in the NICO
dataset. Evidently, the background contexts of the images (e.g sky, road,
grass) highly contribute to ERM's classification process, in contrast to our
model.  This trend also holds when visualizing saliency maps for images in the
PACS dataset, as depicted in Fig \ref{fig:saliency_pacs}. It should be noted, that each image is hidden from the respective model during training. Our method
consistently classifies images based on pixels present in the object and is
less influenced by pixels corresponding to domain-relevant image attributes. Interestingly enough, in the difficult \emph{Sketch} domain, as well as in the \emph{Photo} domain, our model emphasizes solely on the pixels constrained by and adjacent to the sketch and object, while paying less to none attention to the background. This visual evidence seems to confirm our intuition on the learning process of our framework.

\section{Conclusions}

In this work, we introduce $\mathbf{M^2}$; an effective and highly adaptable approach to image classification in the Domain Generalization setting. The main idea behind our approach was to process potentially important information throughout a CNN by taking advantage of multi-scaled outputs from intermediate layers. To this end, we introduce an extraction block consisting of multiple parallel concentration pipelines which enables the extraction of multi-scale disentangled representations and causal features of a class. Our approach was empirically validated through multiple experiments as it was able to achieve top results in established benchmarks. To further push the performance of our initial model and surpass the state-of-the-art, we devised a novel training objective ($\mathbf{M^2}$\textbf{-CL}) tailored to our method, which enables the model to maximize its class-discrimination ability by producing similar representations along samples of the same class. Further experiments on four publicly available benchmark datasets show that our method achieves state-of-the-art results in most of the cases, while remaining highly competitive in the remaining settings. Through ablation studies and visual examples, we demonstrate the necessity of each component in our framework and its capability of disentangling the features of an object and making
predictions based solely on its causal characteristics. Our method however, is not without drawbacks. It's main issue is the additional memory overhead added by the concatenated feature maps before the classification head of our model. In addition, similarly to most contrastive losses, $\mathbf{M^2}$\textbf{-CL} expects a large batch size to function as expected, adding an additional constraint to model training. For future work, we aim to work past the above limitations and perhaps incorporate causal inference or attention mechanisms for processing the disentangled representations, as well as researching additional similarity metrics such as KL Divergence for the distribution of the extracted representations.


\bibliographystyle{IEEEtran}
\bibliography{egbib}

\begin{IEEEbiography}[{\includegraphics[width=1in,height=1.25in,clip,keepaspectratio]{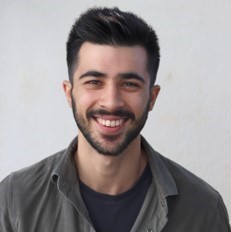}}]
	{Aristotelis Ballas} is currently working toward the Ph.D. degree in computer science with the Department of Informatics and Telematics, Harokopio University of Athens, Greece. His research interests include machine learning and representation learning, with an emphasis on domain generalization and AI in healthcare.
\end{IEEEbiography}

\begin{IEEEbiography}[{\includegraphics[width=1in,height=1.25in,clip,keepaspectratio]{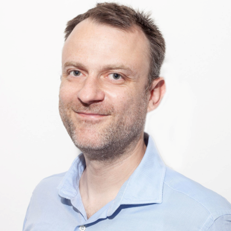}}]{Christos
		Diou} \textit{Member}, IEEE, is an Assistant Professor of Artificial Intelligence and Machine
	Learning at the Department of Informatics and Telematics, Harokopio University
	of Athens. He received his Diploma in Electrical and Computer Engineering and
	his PhD in Analysis of Multimedia with Machine Learning from the Aristotle
	University of Thessaloniki. He has co-authored over 80 publications in
	international scientific journals and conferences and is the co-inventor in 1
	patent. His recent research interests include robust machine learning
	algorithms that generalize well, the interpretability of machine learning
	models, as well as the development of machine learning models for the
	estimation of causal effects from observational data. He has over 15 years of
	experience participating and leading European and national research projects,
	focusing on applications of artificial intelligence in healthcare.
\end{IEEEbiography}

\end{document}